\documentclass[conference]{IEEEtran}
\usepackage{cite}
\usepackage{amsmath,amssymb,amsfonts}
\usepackage{algorithmic}
\usepackage{graphicx}
\usepackage{textcomp}
\usepackage{xcolor}
\def\BibTeX{{\rm B\kern-.05em{\sc i\kern-.025em b}\kern-.08em
    T\kern-.1667em\lower.7ex\hbox{E}\kern-.125emX}}
\graphicspath{{figures/}}
\begin{document}

\title{Machine Learning Systems in the IoT:\\Trustworthiness Trade-offs for Edge Intelligence}

\author{\IEEEauthorblockN{Wiebke Toussaint}
\IEEEauthorblockA{\textit{Engineering Systems and Services} \\
\textit{Delft University of Technology}\\
Delft, Netherlands \\
w.toussaint@tudelft.nl}
\and
\IEEEauthorblockN{Aaron Yi Ding}
\IEEEauthorblockA{\textit{Engineering Systems and Services} \\
\textit{Delft University of Technology}\\
Delft, Netherlands \\
aaron.ding@tudelft.nl}
}

\maketitle

\begin{abstract}
Machine learning systems (MLSys) are emerging in the Internet of Things (IoT) to provision edge intelligence, which is paving our way towards the vision of ubiquitous intelligence. However, despite the maturity of machine learning systems and the IoT, we are facing severe challenges when integrating MLSys and IoT in practical context. For instance, many machine learning systems have been developed for large-scale production (e.g., cloud environments), but IoT introduces additional demands due to heterogeneous and resource-constrained devices and decentralized operation environment. To shed light on this convergence of MLSys and IoT, this paper analyzes the trade-offs by covering the latest developments (up to 2020) on scaling and distributing ML across cloud, edge, and IoT devices. We position machine learning systems as a component of the IoT, and edge intelligence as a socio-technical system. On the challenges of designing trustworthy edge intelligence, we advocate a holistic design approach that takes multi-stakeholder concerns, design requirements and trade-offs into consideration, and highlight the future research opportunities in edge intelligence.

\end{abstract}

\begin{IEEEkeywords}
edge intelligence, machine learning systems, Internet of Things, trade-offs, trustworthiness, smart services
\end{IEEEkeywords}

\section{Introduction}

Machine learning systems are omnipresent and tireless silent helpers that bring order to our busy modern life: they guide us through traffic, classify and predict diseases in humans and plants, and are our eyes and ears in situations where we cannot see and hear. Their underlying machinery are machine learning algorithms that fit complex functions over data to discover patterns and correlations which can be exploited to discover trends and relationships, and for making predictions~\cite{Hastie2009}. Many machine learning algorithms can be scaled to very large datasets and improve with more data. This has made them extremely successful in analysing the large volumes of data produced by digital, online services and applications. Deep neural networks in particular have produced state of the art results for many perception based tasks and are now widely used to process image, video, speech, audio and sequential data~\cite{Lecun2015}. Machine learning is a promising technique when a system or process is not well understood, or too complex and difficult to model explicitly, but data that can surface insights about it has been collected~\cite{mitchell2006}. Equally, if applications are dynamic and evolve over time, machine learning systems can use new data to discover patterns and update their predictions, thus adapting with the application. 

The Internet of Things (IoT) \cite{minerva2015towards} has matured from a vision of digitally connected devices to one of smart services~\cite{georgakopoulos2020internet} and ubiquitous intelligence. For example, a security camera that streams video footage to a remote server is no longer sufficient. Instead, the camera is expected to provide a smart service, such as counting people, or detecting an intruder, thus becoming an intelligent system rather than a mere device connected to the Internet. An intelligent system in the IoT distinguishes itself by having data processing capabilities \cite{Ibarra2017}, meaning that raw sensor data can be transformed to information and knowledge. This kind of abstraction allows humans to infer actionable insights about the system, which can be used to create services that add value to society~\cite{bouguettaya2017service}. 

Historically, human cognition has been needed to abstract information and knowledge from data. However, with the success of machine learning algorithms, new types of technology-driven intelligent systems are emerging that can deliver smart services with reduced human intervention. Machine learning systems are widely investigated to process sensor data and manage system performance and operation in the IoT \cite{Samie2019}. They can be viewed from two perspectives: machine learning systems \textit{for} the IoT support system management and organisation. These systems are designed in service of the IoT and use machine learning to improve overall system aspects like security \cite{algaradi2020survey}, network traffic profiling and IoT device identification~\cite{cui2018survey}. We focus on a second perspective in which machine learning systems are viewed as technical \textit{components of} the IoT that perform advanced data processing tasks like activity recognition, object identification or keyword detection, in service of the greater application objective. We refer to this perspective as edge intelligence.

This paper motivates for an interdisciplinary approach that considers multi-stakeholder concerns, design requirements and trade-offs to develop trustworthy edge intelligence for smart services. In classical machine learning these are not considered, as reliable, abundant, scalable and almost-free communication networks and computing power under control and ownership of a single entity are assumed. In Section~2 we present an overview of machine learning systems and current concerns arising due to training, data, inference and operations. Section~3 highlights additional challenges that the IoT imposes on machine learning systems. In Section~4 we consider edge intelligence trade-offs from a socio-technical and multi-stakeholder perspective. Section~5 presents an outlook for trustworthy edge intelligence, where we take the concerns arising both in machine learning systems and the IoT into consideration. We then highlight opportunities for trustworthy edge intelligence and finally conclude in Section~6.
\section{Overview of Machine Learning Systems}

Machine learning algorithms learn models from data by approximating useful functions that transform input variables, or features, to an output. This is called model training. Trained models are used to calculate an output value for a new input value, which is called inference~\cite{Hastie2009}. When ground truth values, or labels, of the output values are available and used for training a model, the process is called supervised learning. A typical machine learning workflow involves data processing, model training and validation, and inference steps. Data processing serves two purposes. Firstly, input data is cleaned by removing outliers, missing values and errors. Secondly, data is transformed, for example by filtering the features that are used in the learning process. Machine learning systems must facilitate the ongoing deployment of machine learning workflows, which requires that they take operational aspects into consideration. This section highlights data, training, inference and operational concerns that currently challenge machine learning systems. Rather than being exhaustive, we aim to raise important considerations that are bound to impact supervised machine learning systems used in the IoT to provision smart services.

\subsection{Model Training Concerns}

Supervised model training is an iterative optimization problem that aims to find generalizable patterns in data. In statistical learning, the goal of model training is to find the candidate model from limited training data, that has the best predictive performance on new data~\cite{bishop2006pattern}. Practically, the training process determines the values of those model parameters that minimize the error between a predicted value and its corresponding real value in the training data. In addition to parameter coefficients, a model can also have hyperparameters that control its complexity. To find the best model, a range of different model types, parameters and hyperparameters must be explored so that the best model can be selected. However, exploring each of these choices requires computational power, time and energy, resulting in trade-offs between predictive performance and resource consumption. State-of-the-art machine learning models, in particular deep neural networks, can have millions of parameters. Training them takes weeks or even months, and the computing and energy resources required are substantial. Two important approaches for improving the performance of machine learning systems are designing them together with specialized hardware and distributing model training for massive, parallel deployment across cloud servers \cite{Verbraeken2019}. 

In addition to the resources consumed during training, supervised machine learning requires labelled training data, which can be expensive and time consuming to collect~\cite{Ratner2016}. When this is not possible, unsupervised learning which requires no ground truth labels, weak supervision with automated label generation \cite{Ratner2016}, and approaches that reduce the amount of labels required~\cite{Renggli2019} can be considered. Generally these present trade-offs against predictive performance. Due to the infrastructure requirements and cost of model training and data labelling, many applications download pre-trained models from online repositories, which can sometimes be used off-the-shelf, or otherwise adapted to new domains or datasets with transfer learning \cite{pan2010survey}. While pre-trained models speed up the development of new applications, they present significant security risks~\cite{ji2018modelreuse}.

Classical machine learning algorithms are developed under static, benign, closed-world assumptions: they assume that the world does not change, that the environment is good-natured \cite{goodfellow2018making}, and that all categories to be predicted were known during training and contained in the training data \cite{fei2016learning}. Obviously this does not correspond with reality. For example, in medical image classification, it has been established that training data can contain unrecognised categories that are not in the labels but that affect predictive outcomes~\cite{oakdenrayner2020hidden}. While adversarial machine learning \cite{huang2011adversarial} can be used to improve the robustness of models under the malicious attack of an adversary,  and lifelong learning \cite{chen2016lifelong} provides methods for continuous learning by accumulating and maintaining knowledge which can be used to improve future learning, the conditions under which different paradigms can be combined, and what vulnerabilities this may result in, are not obvious. Despite the success of machine learning algorithms, many challenges thus remain to train models that generalize well and have good predictive accuracy while also being resource-efficient, robust, and adaptive in new, real-world environments.

\subsection{Data Provenance Concerns}
The quality of a machine learning model is strongly influenced by the quality and underlying distribution of the data that was used to train it~\cite{Domingos2012}. Due to this central role of data in machine learning, common features of raw observational data, like missing values, data redundancy and noise, significantly impact the performance of the model that is trained. Noise, for example, obscures the data signal and can result from random or systematic errors in the observations, or from data that has been tampered with. Model performance can be degraded further by propagating data errors that were generated during data processing through the entire machine learning workflow. To extend a software metaphor, such data errors are to machine learning systems what bugs are to code~\cite{Breck2019}. Once deployed, data discovery and management are a particular challenge. Datasets are often taken from different sources. As projects grow, so do dependencies between datasets. Over time the training data becomes increasingly complex to track and version~\cite{Amershi2019}. Data dependencies and feedback loops are often hidden and can have unexpected effects that make machine learning systems brittle and error diagnosis expensive~\cite{Sculley2015}. Machine learning systems are also vulnerable to attacks that exploit their dependency on data by polluting training data (poisoning) or modifying input data before inference (evasion)~\cite{ji2018modelreuse}.

\subsection{Inference Concerns}
Trained models infer output values for new data inputs to inform decisions or take actions on an ongoing basis. A trained model is a reusable asset that will make thousands or even millions of predictions before it is retrained.  Unlike model training which happens in the background of an application, inference usually serves users directly and consequently needs to be efficient, reliable and interpretable. Even though the resource requirements for a single prediction are negligible in comparison to those of training a model, the scale at which inference happens requires efficient and optimized processes with high throughput, low latency and graceful performance degradation~\cite{lee2018pretzel}. Traditionally, more attention has been devoted to optimizing the training process rather than inference. Recent releases of popular machine learning platforms like TensorFlow and MXnet now offer libraries for model optimization, but efficiency alone is not enough. When machine learning systems make decisions and act on our behalf, inference must also be reliable \cite{Stoica2017}. Current machine learning systems do not offer predictable throughput, latency and accuracy. Methods that guarantee model outputs and offer reliable uncertainty estimates are needed to provide inference with quality assurance ~\cite{Abdelzaher2020}. Additionally, interpretable inference, which can be likened to the ability of humans to understand how a model works, is necessary for trusted, fair and ethical decision-making based on predictions~\cite{Lipton2018}.

\subsection{Machine Learning Operations (ML Ops) Concerns}
The code responsible for model training and inference is only a small component of the greater system, which includes components for configuration, data collection, data verification, feature extraction, machine resource management, analysis, process management, serving infrastructure and monitoring. Even though machine learning systems are constructed from these different components, models are not modular in the way that software is~\cite{Sculley2015}. Model parameters are learned iteratively, and as dependent on the data distribution as on the features used for training and the hyperparameters. Due to these dependencies, individual models are not extensible and multiple models interact in non-obvious ways. However, models evolve as data changes, methods improve or software dependencies change \cite{Schelter2018}. Ongoing deployment, customisation, reuse and tracking are thus continuous challenges. Machine learning systems require end-to-end software support that facilitates the development, testing, configuration, deployment, management and maintenance of all components that affect data provenance, model training and inference \cite{Ratner2019a}.
\section{IoT Challenges for Machine Learning Systems}

Machine learning systems in smart services are constrained by the nature and requirements of the IoT: distributed, physically-bounded and resource-limited, wireless-connected computing devices that must deliver dynamic and context-aware functionality over multi-layered, heterogeneous architectures \cite{bouguettaya2017service}. For machine learning systems this is both an opportunity and a challenge. By learning from data, they are well suited to offer IoT applications functionality that enables them to adapt to specific locations, environmental or social situations and to evolve with them over time. It could even be argued that machine learning systems are a prerequisite for delivering smart services at scale, as explicitly and perpetually defining and programming the logic for the IoT and its interactions with the physical world and social systems is impossible. At present, however, machine learning systems assume homogeneous and context-independent cloud computing infrastructure with scalable data processing and storage, uninterrupted and unrestricted power supply, and low latency and high bandwidth networks. This stable and consistent environment does not exist for the billions of connected devices in the IoT, where data offloading to wireless networks, distributed, heterogeneous computing infrastructure and resource-constrained devices with physical hardware limitations present trade-offs against each other and the performance of algorithms. To achieve scale, components in the IoT must also be reusable.

\subsection{Offloading to Wireless Communication Networks}
Wireless communication networks like Bluetooth and WiFi connect devices either as a local network, or they connect individual devices and local networks to the Internet \cite{Samie2016}. Many IoT applications rely on wireless connections to offload data collected by devices to the cloud, where it can be cleaned and fused with other datasets, machine learning models can be trained, inference can be done and the data is stored for future use. Offloading gives access to greater computing power and storage, but poses privacy and performance concerns. Wireless communication links have a fixed throughput capacity and range~\cite{Samie2016}, are lossy and noisy~\cite{AlFuqaha2015}, and expose new attack surfaces~\cite{sen2018trifecta}. Network interruptions are bound to affect IoT applications. At worst, machine learning systems must consider the risk of completely loosing connectivity during training or inference, making fault tolerance a necessary consideration~\cite{Abdelzaher2020}. At best, wireless connections introduce latency, variability, uncertainty and costs to machine learning systems, which historically have abstracted away their iterative communication requirements. Offloading thus weighs against privacy and real-time inference requirements, and constrains the frequency, size and data distribution of training updates of machine learning systems. While the data path, timing and transfer volumes can be optimized through routing schemes, scheduling and data compression to minimize bottlenecks and communication costs \cite{Luo2006}, this can reduce predictive accuracy and may be limited by the power supply and computing capabilities of devices~\cite{Samie2016}. 

\subsection{Distribution Across Heterogeneous Devices}
IoT endpoints (e.g. servers, sensors or mobile phones) that are located at the periphery of the Internet are called the edge. Edge computing extends the computing power of the cloud to the endpoints \cite{Sitton2019}, thus creating a geographically distributed network of processors for model training and inference. The edge varies in computing capabilities and connectivity from sensing and actuator devices that observe and control the environment at the lowest level, to gateways and cloud servers. Data processing, model training and inference on the edge can be device, gateway or cloud-centric \cite{Samie2016}. Device-centric approaches reduce offloading challenges, but processing is limited by the computing capabilities and power supply of devices. Gateway-centric computation requires wireless communication, and introduces associated variability and uncertainty. Cloud-centric approaches offer unlimited storage and data processing capabilities, but come with copious communication overheads. Edge servers present an intermediate solution that offers stable power supply and processing closer to the points of data collection, while reducing the data transfer requirements that would be required by the cloud. A simple heuristic is that the availability of data processing, memory, storage and communication overheads all rise with increasing distance from devices. Increasing the former is desirable, while increasing communication overheads is not. The key challenge of distributing machine learning systems in the IoT is to decide whether, when and how to offload computations; that is, to find the optimal balance between local processing and computation offloading given unpredictable networks, and constrained and diverse devices and servers. 

\subsection{Resource Limited Devices}
\label{sec:ondeviceml}
Battery-powered IoT devices have limited memory, processing and power supply and the resources that are available are shared between data collection, data processing (e.g. error detection, compression and encryption) and communication tasks~\cite{Samie2016}. Despite these limitations, on-device machine learning aims to do inference, partial model training and retraining locally on the device to remove the constraints associated with wireless communication. To make machine learning tasks in such resource constrained settings feasible, energy efficiency is of the essence~\cite{Farella2017}. The key requirements for this are to reduce the model size, the energy consumption and the processing requirements of model training and inference, while providing comparable predictive accuracy to what can be achieved on the cloud \cite{Banbury2020}. For mobile devices, federated learning~\cite{BrendanMcMahan2017} has become the standard for distributing model training. This approach reduces privacy concerns and data transfer volumes by processing sensitive data on devices and only performing global parameter aggregation in the cloud. Extensions to federated learning add differential privacy \cite{McMahan2018} to provide privacy guarantees. In federated learning systems for resource-constrained IoT networks, data transfer volumes, model training time and the temperature of devices present trade-offs~\cite{feraudo2020colearn}. 

Classical deep learning models can be several gigabits large. Small models are necessary for on-device inference for two reasons: on-device storage is low, and inference with larger models requires more computations, which consume more energy. To reduce the model size, quantization and pruning are used for model compression~\cite{Iandola2016}. Quantization, which reduces the floating point precision of parameters and gradients, can be rule-based~\cite{Rusci2020} or automated~\cite{Wang2019}, with mixed bitwidths or optimized single bitwidth~\cite{chin2020one}. On the extreme end, binarized neural networks are quantized to 1, 2 or 3 bits~\cite{Fromm2020} and provide superior efficiency, but at the cost of predictive accuracy. Mappings of binary neural networks to look-up tables on Field Programmable Gateway Arrays are able to reduce the energy consumption and latency even more~\cite{Chidambaram2020}. Model pruning eliminates insignificant parameters from neural networks to reduce their size. Despite its popularity, advances in and the impact of model pruning are difficult to evaluate, as the field lacks standardized performance benchmarks~\cite{Blalock2020}. A rising trend for on-device deep learning is the co-design of model and hardware architectures~\cite{stamoulis2019singlepath}, and the exploration of a large search space of possible architectures with Neural Architecture Search~\cite{Zoph2017}.

\subsection{Component Reusability}

Reusability is an important design consideration in the IoT and
a necessity for deploying smart services at massive scale~\cite{bouguettaya2017service}. This means that IoT components must be discoverable and useable by third parties to deploy new services. Components that lend themselves to reuse are hardware, data, models and the execution environments. For model training, raw data, features, sensing and processing devices can be shared. Similarly, for inference the sensors and processors, observational and transformed data streams, and models can be shared. Shared devices reduce the cost of hardware acquisition and system life-cycle cost (e.g. maintenance activities), which is an advantage. However, shared components bring their own challenges. Shared hardware and models challenge machine learning systems to consider hardware heterogeneity and utilization, workload allocation and prioritization, process scheduling and isolation, resource management and security. When many devices operate in close proximity, interference can affect data transmission, leading to increased energy consumption of devices, reduced service quality and communication delays. Shared data additionally poses questions of anonymity and control, governance and persistence: for example, who grants access to your phone's geolocation data to track your digital footsteps through the city? Do those that see your trail know it's you? And are you able to wipe your trace when you want to?  

Sensing devices and smart services can be mapped in one-to-one, one-to-many, many-to-one and many-to-many configurations~\cite{Samie2016}. Training and inference workloads can be mapped to processing devices in a similar fashion. Collaborative inference with data inputs from multiple sensing devices, and multi-tenant processing which allocates and schedules multiple workloads over one or more resources, are the logical extension of pervasive sensing and edge intelligence to ubiquitous intelligence. Distributed machine learning operations for edge intelligence are bound to be complex and complicated. The heterogeneous and geographically dispersed IoT will amplify the operational challenges already observed in the cloud. Moreover, sharing presupposes the involvement of multiple stakeholders, which inherently implies that ownership, governance, accountability and trust matter.
\section{Multi-Stakeholder Trade-Offs}

The IoT is not only a complex collection of technologies, but a socio-technical system in a multi-stakeholder environment \cite{nist2017framework} with networks of independent actors consisting of users, data generators, network providers, data processors, application service providers and many others. With so many players involved, data and device use, management, maintenance and ownership are heterogeneous and can change. This multi-stakeholder environment gives rise to conflicting requirements and priorities between actors that must be considered when designing edge intelligence for smart services.

\subsection{Design Aspects and Stakeholder Concerns in the IoT}

Engineered systems are designed to deliver reliable, predictable and robust performance within acceptable bounds of confidence, in an unpredictable world. For example, boarding a plane when a thunderstorm is brewing, you have confidence that you will arrive at your destination because you have a justified belief that the plane was carefully designed, that it is operated by a well trained pilot and that the air traffic control system abides by internationally regulated standards of excellence. In its vision of smart services and ubiquitous intelligence, the IoT\footnote{The definitions of the IoT and cyber physical systems (CPS) have been converging over time \cite{Greer2019}. We take a unified perspective of the two fields and refer to them collectively as IoT, to retain focus on machine learning systems.} serves as subsystem to larger, yet again socio-technical, engineered systems. Its hybrid cyber-physical nature however means that actions in the cyber realm carry consequences in the physical environment and can influence our experience of the world, like getting cold when a heating system is deactivated. This imposes more stringent requirements on its design than what would be the case for purely physical or solely cyber systems. 

Specifications for the IoT are captured in standards (e.g. see references listed in \cite{nist2017framework}). A useful approach for identifying system requirements is through \textit{concerns} that are of interest to one or more stakeholders~\cite{nist2017framework}. Table~\ref{tab:aspects_concerns} lists concerns, grouped into \textit{aspects} based on common attributes, that have been developed to provide a comprehensive framework for the design of hybrid cyber and physical systems, like the IoT. Concerns are related and composable. For example, in considering the uncertainty concern, the latency imposed by specifying and managing uncertainty must also be considered. Typically concerns present trade-offs and stakeholders are likely to prioritize them differently. Requirements can be used to express system properties that address relevant concerns.

\begin{table}[ht]
\begin{center}
\caption{Aspects and concerns of IoT/CPS \cite{nist2017framework}}
\label{tab:aspects_concerns}
\small
\begin{tabular}{lp{0.65\linewidth}} \smallskip 
    \textbf{Aspects} & \textbf{Concerns} \\\hline\noalign{\medskip}\smallskip
    \textbf{functional} & actuation, communication, controllability, functionality, manageability, monitoriablity, performance, physical, physical context, sensing, states, uncertainty \\ \smallskip
    \textbf{business} & enterprise, cost, environment, policy, quality, regulatory, time to market, utility \\ \smallskip
    \textbf{human} & human factors, usability \\ \smallskip
    \textbf{trustworthiness} & privacy, reliability, resilience, safety, security \\ \smallskip
    \textbf{timing} & logical time, synchronization, time awareness, time-interval and latency \\ \smallskip
    \textbf{data} & data semantics, identity, operations on data, relationship between data, data velocity, data volume \\ \smallskip
    \textbf{boundaries} & behavioural, networkability, responsibility\\ \smallskip
    \textbf{composition} & adaptability, complexity, constructivity, discoverability\\ \smallskip
    \textbf{lifecycle} & deployability, disposability, engineerability, maintainability, operability, procurability, producibility\\
\end{tabular}
\end{center}
\vspace{-2.5em}
\end{table}

\subsection{Implications for the Design of Edge Intelligence}

Edge intelligence integrates machine learning systems into the cyber system of the IoT. Unlike the low risk analytical settings in which statistical machine learning has been developed, this can have real-world, potentially harmful or even life-threatening repercussions if the system malfunctions or fails. As a component of the IoT, it is thus necessary that machine learning systems for edge intelligence conform to the requirements of the IoT. And as with other software systems, specifying the target system behaviour during a requirements analysis process is essential. Machine learning systems in the wearables domain already incorporate explicit requirements analysis processes to specify system requirements upfront \cite{Fortino2013}. This is not the norm in other domains, and the opportunity exists to develop approaches for navigating conflicting design concerns and requirements trade-offs. These will need to consider the multi-layered and complex component technologies for edge intelligence, the limitations that they present individually and collectively, and the design choices that satisfy the prioritized requirements of stakeholders.

\noindent \textbf{Opportunity: }\textit{Frameworks and processes are needed to elicit stakeholder requirements, navigate conflicting design concerns and prioritize trade-offs to make informed design choices for edge intelligence.}

True to its statistical heritage, the (implicit) design of machine learning systems in the IoT focuses primarily on feature engineering, algorithm selection, parameter optimization and architecture design, with the goal of optimizing predictive performance. From an IoT perspective, this addresses the performance concern of the functional aspect, but falls short on measuring and optimizing for other concerns. On-device machine learning (see Section \ref{sec:ondeviceml}) already broadens concerns to account for physical contexts with resource limitations. Likewise, wireless offloading raises uncertainty, privacy, security, latency, data velocity and volume concerns, while distribution and device heterogeneity introduce controllability and synchronization concerns. Within a service paradigm, quality plays an important role, as it is viewed as a discriminating factor by which users choose services \cite{bouguettaya2017service}. Providing ways for estimating uncertainty and for measuring, controlling and guaranteeing quality of service thus carry particular significance for smart services.

\noindent \textbf{Opportunity: }\textit{Metrics and benchmarks beyond predictive performance are needed so that machine learning systems for edge intelligence can be specified, designed and evaluated.}

Issues of fairness, accountability and transparency are endemic to machine learning systems~\cite{Wallach2014}, where the data quality and distribution is integral to the model that is learnt. Models learned from data have the unfortunate drawback that they propagate the biases of the data collection process. Moreover, some machine learning algorithms, like deep neural networks, are considered to be "black box" algorithms, meaning that the inner workings of the algorithm according to which predictions are made are poorly understood and not controllable by humans. At present, IoT concerns do not consider concerns such as fairness, transparency, explainability and interpretability, that arise due to the data-centric nature of machine learning systems. They need to be accounted for to avoid becoming a blind spot in the design of edge intelligence.

\noindent \textbf{Opportunity: }\textit{To be relevant to edge intelligence, concerns and aspects of the IoT need to be expanded to incorporate well known challenges due to the data-centric nature of machine learning systems.}
\section{Outlook: Trustworthy Edge Intelligence}

Trust-in-technology research extends trust beyond social systems to non-human, artificial entities. Technologies vary in their perceived "humanness", and users trust technologies differently based on this~\cite{lankton2015technology}. If the perceived humanness of a technology is high, then human-like trust constructs such as benevolence, integrity and ability, are good measures of trust. Congruently, if the perceived humanness is low, then system-like trust constructs like helpfulness, reliability and functionality are more appropriate measures. While related, trust and trustworthiness represent different concepts \cite{cho2011perceived}. Trust is a psychological state that indicates whether a trustor is willing to take risks for a trustee in the absence of monitoring or external control. Trustworthiness is a necessary condition for choosing to trust someone and focuses on the characteristics of a trustee. Trustworthiness concerns are essential considerations in both AI and the IoT, but they are approached from different perspectives in the two fields.

\subsection{Trustworthiness in AI}

Technologies that create the perception of social presence of other humans, that facilitate social behaviour (e.g. engaging in dialogue or receiving affection), and that enable interactions with other people are perceived to be more human-like~\cite{lankton2015technology}. Artificial intelligence technologies, which encompass machine learning systems, are thus human-like by definition and design. Heightening public mistrust has lead governments and organisations to rapidly develop AI frameworks to specify principles for trustworthy AI. The core themes that emerge from prominent frameworks are: privacy, accountability, safety and security, transparency and explainability, fairness and non-discrimination, human control of technology, professional responsibility, and promotion of human values~\cite{fjeld2020principled}. These trust constructs resonate with the perspective that AI technologies are perceived to be human-like. 
While the frameworks lay the theoretical ground work, to be useful trustworthy AI needs to develop measurable trustworthiness concerns that can lead to practical and enforceable specifications.  

\noindent \textbf{Opportunity: }\textit{Trustworthiness concerns of machine learning systems need to be standardized and operationalized so that they can be incorporated in specifications and evaluated objectively in applications.}

\subsection{Trustworthiness in the IoT}

In contrast, trustworthiness concerns in the IoT are agreed on across the industry, captured in standards, and formally defined as safety, security, privacy, reliability and resilience~\cite{nist2017framework}. The concerns serve to assure that systems behave as expected under various operating conditions. They support the view that the IoT is perceived to be less human-like, and more system-like. Other properties such as controllability, manageability, functionality, performance and uncertainty are considered as functionality concerns, rather than trustworthiness concerns.

\subsection{Opportunities for Trustworthy Edge Intelligence}

Neither trustworthy AI, nor trustworthiness concerns in the IoT address the full spectrum of trustworthiness concerns that arise in edge intelligence. For example, a machine learning system may fail to make correct predictions under open world assumptions, which can include new categories, unseen examples, black swan events and foreign attack models. To be able to perform fault diagnosis in such scenarios, explainability is a necessary requirement. Or consider a smart camera installed in a new context where the population does not resemble the people that were represented in the training data of the deployed model. The machine learning system may fail to recognize members of that population and the trust constructs of fairness and non-discrimination will directly impact the functionality of the application. On the other hand, intermittent and unreliable data transfer over wireless channels can result in missing values that limit inference quality and affect system level predictive performance. A voice assistant that alerts emergency response when you cry for help, will need to perform reliably even in those settings. Trustworthy edge intelligence thus requires that trust constructs for machine learning systems and trustworthiness concerns arising in the IoT are considered together. As with other design requirements, trustworthiness concerns will be composable and pose trade-offs against each other and against other stakeholder concerns. There is thus a need to:

\begin{itemize}
    \item analyze the trustworthiness concerns that arise in machine learning systems for edge intelligence and smart services
    \item explore the overlap and trade-offs of trustworthiness concerns between machine learning and IoT systems
    \item characterize the interactions and trade-offs between trustworthiness concerns and other stakeholder concerns
    \item expand research into trustworthy machine learning to also address the diverse spectrum of challenges and trade-offs that arise in edge intelligence
\end{itemize}

\section{Concluding Remarks}

Ever-growing, densely populated urban centers need to monitor, track, care for and nurture their social, natural and artificial systems. Smart services, informed by ubiquitous intelligence, are viewed as a way of doing this. Machine learning systems can enable smart services by provisioning the IoT with edge intelligence, giving rise to ubiquitous intelligence. This paper presents challenges and trade-offs that arise when designing trustworthy edge intelligence for smart services. Despite the maturity of machine learning systems and the IoT, combining the two technologies presents new concerns for edge intelligence. One the one hand, many machine learning systems have been deployed in large-scale production environments, and the model training, data provenance, inference and ongoing operational challenges are known. These challenges prevail when deploying machine learning systems in the IoT, but are not considered in existing IoT design frameworks. On the other hand, additional challenges arise due to communication offloading, distributed, heterogeneous and resource-constrained devices, and the need to share and reuse components in the IoT. These challenges are not addressed by classical machine learning, or large scale, cloud-based machine learning systems.

We position machine learning systems as a component of the IoT, and edge intelligence as a socio-technical system. We motivate that multi-stakeholder concerns, design requirements and technology trade-offs should be taken into consideration when developing edge intelligence, and highlight opportunities that exist to facilitate this. With an outlook on trustworthiness, we demonstrate that an interdisciplinary perspective is essential, as trust constructs are considered differently in machine learning systems and the IoT. By combining perspectives, and taking multi-stakeholder concerns, design requirements and trade-offs into considerations, it is possible to perceive of a future where holistic, trustworthy edge intelligence and smart services are possible.

\bibliographystyle{IEEEtran}
\bibliography{library}

\end{document}